\pdfoutput=1

\documentclass[11pt,a4paper]{article}
\usepackage[hyperref]{acl2020}
\usepackage{times}
\usepackage{latexsym}

\usepackage[utf8]{inputenc}
\usepackage{amsmath}
\usepackage{amssymb}
\usepackage{amsthm}
\usepackage{mathtools}
\usepackage{qtree}

\usepackage{microtype}

\aclfinalcopy 

\newtheorem{theorem}{Theorem}[section]
\newtheorem{definition}{Definition}

\newtheorem{lemma}[theorem]{Lemma}

\newtheorem{manualtheoreminner}{Theorem}
\newenvironment{manualtheorem}[1]{%
  \renewcommand\themanualtheoreminner{#1}%
  \manualtheoreminner
}{\endmanualtheoreminner}

\newtheorem{manuallemmainner}{Lemma}
\newenvironment{manuallemma}[1]{%
  \renewcommand\themanuallemmainner{#1}%
  \manuallemmainner
}{\endmanuallemmainner}

\newcommand{\R}{\mathbb{R}}
\newcommand{\N}{\mathbb{N}}

\renewcommand{\S}{\mathbb{S}}
\newcommand{\0}{\mathbf{0}}
\newcommand{\1}{\mathbf{1}}
\newcommand{\ra}{\rightarrow} 
\newcommand{\rra}{\Rightarrow}

\newcommand{\rras}{\xRightarrow{*}}
\newcommand{\ee}{\mathbf{e}}
\newcommand{\ff}{\mathbf{f}}
\newcommand{\dd}{\mathbf{d}}

\newcommand{\inst}[1]{\textsuperscript{#1}}

\title{Tensors over Semirings for Latent-Variable Weighted Logic Programs}

\author{
  Esma Balk{\i}r\inst{1} \quad Daniel Gildea\inst{2} \quad Shay B. Cohen\inst{1} \smallskip \\
  \inst{1}ILCC, School of Informatics, University of Edinburgh \\ \inst{2}Department of Computer Science, University of Rochester \smallskip \\
  \texttt{esma.balkir@ed.ac.uk} \\ \texttt{gildea@cs.rochester.edu} \,
  \texttt{scohen@inf.ed.ac.uk}
}

\date{}

\begin{document}
\maketitle
\begin{abstract}
Semiring parsing \citep{Goodman1999SemiringParsing} is an elegant framework for describing parsers by using semiring weighted logic programs.
In this paper we present a generalization of this concept: latent-variable semiring parsing. With our framework, any semiring weighted logic program can be \textit{latentified} by transforming weights from scalar values of a semiring to rank-n arrays, or tensors, of semiring values, allowing the modeling of latent variables within the semiring parsing framework. Semiring is too strong a notion when dealing with tensors, and we have to resort to a weaker structure: a partial semiring.\footnote{Our definition of a partial semiring is slightly different than those in the abstract algebra literature  e.g.~\citet{Steenstrup1985Sum-OrderedSemirings}.} We prove that this generalization preserves all the desired properties of the original semiring framework while strictly increasing its expressiveness.
\end{abstract}

\section{Introduction}

Weighted Logic Programming (WLP) is a declarative approach to specifying and reasoning about dynamic programming algorithms and chart parsers. 
WLP is a generalization of bottom-up logic programming where proofs are assigned weights by combining the weights of the axioms used in the proof, and the weight of a theorem is in turn calculated by combining the weights of all its possible proof paths. The combinatorial nature of this procedure makes weighted logic programs highly suitable for specifying dynamic programming algorithms. In particular, \citet{Goodman1999SemiringParsing} presents an elegant abstraction for specifying and computing parser values based on WLP where the values could be drawn from any complete semiring. This generalizes the case of Boolean decision problems, probabilistic grammars with Viterbi search and other quantities of interest such as the best derivation or the set of all possible derivations. It is then possible to derive a general formulation of inside and outside calculations in a way that is agnostic to the particular semiring chosen.

Latent variable models have been an important component in the NLP toolbox. The central assumption in latent variable models is that the correlations between observed variables in the training data could be explained by unobserved, hidden variables. Latent variables have been used with grammars such as Probabilistic Context-Free Grammars (PCFGs), where each node in the parse tree is represented using a vector of latent state probabilities that further extend the expressiveness of the grammar \citep{Matsuzaki2005ProbabilisticAnnotations}.

The approach of adding latent variables to formal grammars
have proven to be a fruitful one: in the context of PCFG parsing, \citet{Matsuzaki2005ProbabilisticAnnotations} show that latent variable PCFGs (L-PCFGs) perform on par with models hand-annotated with linguistically motivated features. \citet{Cohen2013ExperimentsPCFGs} report that on the Penn Treebank dataset, L-PCFGs trained with either EM or a spectral algorithm provide a 20\% increase in F1 over PCFGs without latent states. \citet{Gebhardt2018GenericParsing} shows that the benefits of latent variables are not limited to PCFGs by successfully enriching both Linear Context-Free Rewriting Systems and Hybrid Grammars with latent variables, and demonstrates their applicability on discontinuous constituent parsing.

Given the usefulness of latent variables, it would be desirable to have a generic inference mechanism for any latent variable grammar. WLPs can represent inference algorithms for probabilistic grammars effectively. However, this does not trivially extend to latent-variable models because latent variables are often represented as vectors, matrices and higher-order tensors, and these taken together no longer form a semiring. This is because in the semiring framework, values for deduction items and for rules must all come from the same set, and the semiring operations must be defined over all pairs of values from this set.  This does not allow for letting different grammar nonterminals be represented by vectors of different sizes.  More importantly, it does not allow for a rule's value to be a tensor whose dimensionality depends on the rule's arity, as is generally the case in latent variable frameworks.

In this paper we start with a broad interpretation of latent variables as tensors over an arbitrary semiring. While a set of tensors over semirings is no longer a semiring, we prove that if the set of tensors have certain matching dimensions for the set of grammar rules they are assigned to, then they fulfill all the desirable properties relevant for the semiring parsing framework. This paves the way to use WLPs with latent variables, naturally improving the expressivity of the statistical model represented by the underlying WLP\@.
Introducing a semiring framework like ours makes it easier to seamlessly incorporate latent variables into any execution model for dynamic programming algorithms (or software such as Dyna, \citealt{Eisner2005CompilingLanguage}, and other Prolog-like/WLP-like solvers).

We focus on CFG parsing, however the same latent variable techniques can be applied to any weighted deduction system, including systems for parsing TAG, CCG and LCFRS, and systems for Machine Translation \cite{Lopez2009TranslationDeduction}.
The methods we present for inside and outside computation can
be used to learn latent refinements of a specified grammar
for any of these tasks with EM \citep{Dempster1977MaximumAlgorithm, Matsuzaki2005ProbabilisticAnnotations}, or used as a backbone to create spectral learning algorithms \citep{Hsu2012AModels, Bailly2009GrammaticalProblem, Cohen2014SpectralComplexity}.

\section{Main Results Takeaway}

We present a strict generalization of semiring weighted logic programming, with a particular focus on parser descriptions in WLP for context-free grammars. Throughout, we utilize the correspondence between axioms and grammar rules, deductive proofs and grammar derivations, and derived theorems and strings. 

We assume that axioms/grammar rules come equipped with weights in the form of tensors over semiring values. The main issue with going from semirings to tensors over semiring values is that these weights  need to be \textit{well defined} in that any valid derivation should correspond to a sequence of well defined semiring operations. For CFGs, we give a straightforward condition that ensures this is the case. This essentially boils down to making sure that each non-terminal corresponds to a fixed vector space dimension.
For example, if $A$ corresponds to a space of $d_1$ dimensions, $B$ to $d_2$ and $C$ to $d_3$, then a rule $A \ra B \,\, C$ would have a tensor weight in $d_2 \times d_3 \times d_1$. 

As long as the weights are well defined, the standard definitions for the value of a grammar derivation and a string according to a semiring weighted grammar extend to the case of tensors of semirings. Weighted logic programming provides the means to declaratively specify an efficient algorithm to obtain these values of interest. In line with \citet{Sikkel1998ParsingAlgorithms} and \citet{Goodman1999SemiringParsing} we present precise conditions for when a partial-semiring WLP describes a correct parser. 





The value of the WLP formulation of parsing algorithms is that it provides a unified fashion in which dynamic programming algorithms can be extracted from the program description. This relies on the ability of a WLP to decompose the value of a proof to a combination of the values of the sub-proofs. Specifically, given a derivation tree, a WLP description automatically provides algorithms for calculating the inside and outside values. We provide analogous algorithms for calculating the inside and outside values for partial-semiring WLPs.
Our outside formulation addresses the non-commutative nature of tensors themselves, and could be extended to cases where the underlying semiring is non-commutative using the techniques presented by \citet{Goodman1998ParsingInside-Out}.


    
    
    
    

\section{Related Work}
``Parsing as deduction'' \citep{Pereira1983ParsingDeduction} is an established framework that allows a number of parsing algorithms to be written as declarative rules and deductive systems \citep{Shieber1995PrinciplesParsing}, and their correctness to be rigorously stated \citep{Sikkel1998ParsingAlgorithms}. \citet{Goodman1999SemiringParsing} has extended the parsing as deduction framework to arbitrary semirings and showed that various different values of interest could be computed using the same algorithm by changing the semiring. This led to the development of Dyna, a toolkit for declaratively specifying weighted logic programs, allowing concise implementation of a number of NLP algorithms \citep{Eisner2005CompilingLanguage}.

The semiring characterization of possible values to assign to WLPs gave rise to the formulation of a number of novel semirings. 
One novel semiring of interest for purposes of learning parameters is the \textit{generalized entropy semiring} \citep{Cohen2008DynamicPrograms} which can be used to calculate the KL-divergence between the distribution of derivations induced by two weighted logic programs. Other two semirings of interest are \textit{expectation} and \textit{variance} semirings introduced by \citet{Eisner2002ParameterTransducers} and \citet{Li2009First-Forests}. These utilize the algebraic structure to efficiently track quantities needed by the expectation-maximization algorithm for parameter estimation. Their framework allows working with parameters in the form of vectors in $\R^n$ for a fixed $n$, coupled with a scalar in $\R_{\geq 0}$. The semiring value of a path is roughly calculated by the multiplication of the scalars and (appropriately weighted) \textit{addition} of the vectors. This is in contrast with our framework where weights could be tensors of arbitrary rank rather than only vectors, and the values of paths are calculated via tensor multiplication.

Finally, \citet{Gimpel2009CubeSemirings} extended the semiring framework to a more general algebraic structure with the purpose of incorporating non-local features. Their extension comes at the cost that the new algebraic structure does not obey all the semiring axioms. Our framework differs from theirs in that under reasonable conditions, tensors of semirings do behave fully like regular semirings. 



\section{Background and Notation}
Our formalism could be used to enrich any WLP that implements a dynamic programming algorithm,  but for simplicity, we follow \citet{Goodman1999SemiringParsing} and focus our presentation on parsers with a context-free backbone.\footnote{Note that given a grammar $G$ in a formalism $F$ and a string $\alpha$, it is possible to construct a CFG grammar $c(G, w)$ from $G$ and $\alpha$ \citep{Nederhof2003WeightedAlgorithm}. This construction is possible even for range concatenation grammars \citep{Boullier2004RangeGrammars} which span all languages that could be parsed in poly-time.}

\subsection{Context-free Grammars}

Formally, a Context-Free Grammar (CFG) is a 4-tuple $\langle N, \Sigma, \mathcal{R}, S \rangle$. The set of $N$ denotes the non-terminals which will be denoted by uppercase letters $A, B$ etc., and $S$ is a non-terminal that is the special start symbol. The set of $\Sigma$ denotes the terminals which will be denoted by lowercase letters $a,b$ etc. $\mathcal{R}$ is the set of rules of the form $A \ra \alpha$ consisting of one non-terminal on the left hand side (lhs), and a string $\alpha \in (N \cup \Sigma)^*$ on the right hand side (rhs). We will use $\alpha \rra \beta$ if $\beta$ could be derived from $\alpha$ with the application of one grammar rule. We will say that a sentence $\sigma \in \Sigma^+$ could be derived from the non-terminal $A$ if $\sigma$ could be generated by starting with $A$ and repeatedly applying rules in $\mathcal{R}$ until the right hand side contains only terminals, and denote this as $A \rras \sigma$. We will denote the language that a grammar $G$ defines by $\mathcal{L}(G) = \{\sigma | S \rras \sigma\}$. 

CFG derivations can naturally be represented as trees. We will use the notation $\langle r: T_1\ldots T_k \rangle$ to represent a tree that has the node $r$ as its root and $T_1,\ldots ,T_k$ as its direct subtrees. We will use $\mathcal{D}_G$ to denote the set of all derivation trees that can be constructed with the grammar $G$, and $\mathcal{D}_G(\sigma)$ for all valid derivation trees that generate the sentence $\sigma$ in $G$.


\subsection{Semirings}
A semiring is an algebraic structure similar to a ring, except that it does not require additive inverses. 

\begin{definition}\label{def-semiring}
A \textbf{semiring} is a set $\S$ together with two operations $+$ and $\times$, where $+$ is commutative, associative and has an identity element 0. The operation of $\times$ is associative, has an identity element 1 and distributes over $+$.
\end{definition}

The set of non-negative integers together with the usual $\times, +, 0, 1$ is a semiring, and so are probability values in $[0,1]$. Booleans \{TRUE, FALSE\} also form a semiring with $\times := \vee$, $+ := \wedge$, $0 :=$ FALSE and $1 :=$ TRUE. 

There are a few less common semirings that provide useful values in parsing.
The \textit{Viterbi} semiring calculates the probability of the best derivation.
It has values in $[0,1]$, $+ :=$ max and $\times, 0, 1$ as standard.
The \textit{Derivation forest}, \textit{Viterbi derivation} and \textit{Viterbi $n$-best} semirings calculate the set of all derivations, the best derivation and the $n$-best derivations respectively. Unlike the previous examples, the $\times$ operation of these semirings is not commutative. In general, if the $\times$ operation in a semiring is commutative, we refer to it as a \textit{commutative semiring}, and otherwise it is referred to as \textit{non-commutative}. For precise definitions and detailed descriptions of these semirings see \citet{Goodman1999SemiringParsing}.

\subsection{Weighted Logic Programming}
A logic program consists of \textit{axioms} and \textit{inference rules} that could be applied iteratively to prove theorems. Inference rules are expressed in the form $\frac{A_1\ldots A_k}{B}$ where $A_1\ldots A_k$ are antecedents from which $B$ can be concluded. Axioms are inference rules with no antecedents.  

One way to express dynamic programming algorithms such as CKY is as logic programs. This approach takes the point of view of \textit{parsing as deduction}: terms consist of grammar rules and \textit{items} in the form of $[i,A,j]$ that correspond to the intermediate entries in the chart. Grammar rules are taken to be axioms, and the description of the parser is given as a set of inference rules. These can have both grammar rules and items as antecedents and an item as the conclusion. A logic program in this form includes a special designated goal item that stands for a successful parse.  

Continuing with the example of CKY, consider the procedural description for how to obtain a chart item from smaller chart items if we have the rule $A \ra B \,\, C$ in the grammar:
\begin{align*}
    chart[i, &A, j] := chart[i, A, j] \; \vee \\ 
    &(chart[i, B, k] \; \wedge \; chart[k, C, j])
\end{align*}
The corresponding inference rule in a logic program would be:
\[\frac{A \ra B\,\,C \;\;\;\;\;\; [i,B,k] \;\;\;\;\;\; [k,C,j]}{[i,A,j]} \]

Note that in the inference rule above, $A \ra B\,\,C$ is a rule template with free variables $A, B, C$. In general, the terms in inference rules can contain free variables, however for a logic program to describe a valid dynamic algorithm, every free variable in the conclusion of an inference rule must appear in its antecedents as well.

A \textit{weighted} logic program is a logic program where terms are  assigned values from a semiring.
When paired with semiring operations, inference rules provide the description of how to compute the value of the conclusion given the values of the antecedents. The result of an application of a particular inference rule is the semiring multiplication of all the antecedents. The value of a term $B$ is then calculated as the semiring sum of values obtained from inference rules that have $B$ as their the conclusion. 
 
 \subsection{Semiring Parsing}
 In the context of parsing, \citet{Goodman1999SemiringParsing} presents a framework where a grammar $G$ comes equipped with a function $w$ that maps each rule in $G$ to a semiring value. Then, a grammar derivation string $E$ consisting of the successive applications of rules $e_1,\ldots ,e_n$ is defined to have the value $V_G(E) = \prod^n_{i=1}w(e_i)$, and the value of a sentence $\sigma \in \mathcal{L}(G)$ is defined as $V_G = \sum^k_{j=1}V_G(E_j)$ where $E_1, E_2,\ldots ,E_k$ are the derivations of $\sigma$ in $G$. 

 A parser specification is given in the form of a weighted logic program, referred to as \textit{item-based description}. From these, the value of a derivation $D$ is calculated recursively as follows:
 \begin{equation*}
     V(D) = 
        \begin{cases}
            w(D) &\text{if } D \text{ is a rule}\\
            \prod_{i=1}^k V(D_i) \hspace{-2.2mm} &\text{if } D = \langle b: D_1,\ldots ,D_k \rangle 
        \end{cases}
\end{equation*}
where $\prod$ is the semiring product. 

Let $inner(x)$ represent the set of all derivation trees headed by the item $x$. Then the value of $x$ is:
\begin{equation*}
    V(x) = \sum_{D \in inner(x)} V(D)
\end{equation*}
where $\sum$ is the semiring addition. The value of a sentence is then equal to $inner(goal)$. 

Given the definitions of value according to the grammar and the parser, \citet{Goodman1999SemiringParsing} provides a theorem for conditions of correctness:
\begin{theorem}\label{thm:correctness-goodman}(\citealt{Goodman1999SemiringParsing}, Theorem 1; informal) An item-based description $I$ is correct if for every grammar $G$ there exists a one-to-one correspondence between the grammar and item derivations, and these derivations get the same value regardless of weight function used. 
\end{theorem}

One caveat with calculating based on item-based derivations is that there is an ordering of items: we cannot compute the value of an item unless the values of all its children are computed already. For this, \citet{Goodman1999SemiringParsing} assumes that each item is assigned to a \textit{bucket} so that if an item $b$ depends on $a$, then $bucket(a) \leq bucket(b)$. If a bucket depends on itself, then it is considered a special \textit{looping} bucket. For all the formulas we present in this the main paper we assume that the items belong to non-looping buckets. The formulas for looping buckets are provided in Appendix B.

For an item $x$, calculating its value might require summing over exponentially many derivation trees. To address this, it is possible to provide a general formula that efficiently computes the inner value for an item (\citealt{Goodman1999SemiringParsing}, Theorem 2): 
\begin{equation*}
    V(x) = \sum_{a_1,\ldots ,a_k \text{s.t.} \frac{a_1,\ldots ,a_k}{x}} \prod_{i=1}^k V(a_i)
\end{equation*}

The other important value associated with an item $x$ is its \textit{outside} value $Z(x)$, which is the sum of values of  derivation trees, modified so that $x$ is removed with all its subtrees. This value is complementary to the inside values $V(x)$ (\citealt{Goodman1999SemiringParsing}, Theorem 4): 
\begin{equation*}
    V(x) \times Z(x) = \sum_{D\,\, \text{a derivation}} V(D) C(D, x)
\end{equation*}
where $C(D, x)$ is the count of the occurrences of item $x$ in derivation $D$.

$Z(x)$ can likewise be calculated using a recursive formula if the values are from a commutative semiring (\citealt{Goodman1999SemiringParsing}, Theorem 5): 
\begin{equation*}
    Z(x) = \hspace{-3mm} \sum_{\substack{j,a_1,\ldots ,a_k,b \;\; \text{s.t.} \\
                            \frac{a_1\ldots a_k}{b} \text{ and } x=a_j}} \hspace{-3mm} Z(b) \times \prod_{i=1}^{j-1} V(a_i) \times \prod_{i=j+1}^k V(a_i)
\end{equation*}

\subsection{Tensor Notation} We use the term \textit{tensor} to refer to an $n$-dimensional array of semiring values. We use $\S$ to denote a semiring and $\mathbf{A}, \mathbf{B}$ etc.\ to denote tensors. The element $\mathbf{A} \in \S^{a_1 \times a_2 \times \ldots  \times a_n}$ will denote that $\mathbf{A}$ is a rank-$n$ tensor of values drawn from $\S$, with the $i$th rank having dimension $a_i$. The entry in index $k_1,\ldots ,k_n$ will be denoted with subscripts $\mathbf{A}_{k_1,\ldots ,k_n}$. 

\section{Latent-variable Parsing as Tensor Weighted Logic Programs}

For semiring parsing to work for latent-variable models it should allow weights to be vectors, matrices and tensors. In this section we present a framework that generalizes that of \citet{Goodman1999SemiringParsing}, and is able to capture tensors over semirings as weights. Note that this includes scalars as a special case.


\subsection{Semiring Operations} 



The main reason why tensors over semirings are not semirings is that with tensor weights, $\oplus$ and $\otimes$ become partially defined -- not all elements can naturally be added or multiplied to any other element anymore. We refer to these structures as \textit{partial semirings}. With some reasonable constraints, we show that $\oplus$ and $\otimes$ obey the semiring axioms in cases that are relevant for the semiring parsing framework. 

Let $\mathbb{S}$ be the chosen underlying semiring, $+, \times$ to be the semiring operations and $\mathbf{0}, \mathbf{1}$ be the additive and multiplicative identity of the semiring respectively. The set of possible weights are defined as $\{\mathbb{S}^{d_1 \times \ldots \times d_n}\}$ for $n \in \N$, and $d_i \in \N$ for all $i \leq n$. $\oplus$ is a partial addition that is defined on two tensors $\mathbf{A}, \mathbf{B} \in \mathbb{S}^{d_1 \times \ldots \times d_n}$ as long as the dimensions of each of their ranks match. Then, the addition is defined component-wise: 
\[ (\mathbf{A} \oplus \mathbf{B})_{i_1,\ldots ,i_n} := \mathbf{A}_{i_1,\ldots ,i_n} + \mathbf{B}_{i_1,\ldots ,i_n} \]

The additive identity is now a class of tensors, one for each unique list of tensor dimensions. The additive identity for any $\mathbf{A} \in \mathbb{S}^{d_1 \times \ldots \times d_n}$ is the tensor $\mathbf{Z} \in \mathbb{S}^{d_1 \times \ldots \times d_n}$ with $\mathbf{0}$ in every entry.

Multiplication is defined as the contraction of an index between two tensors with arbitrary number of ranks. Specifically, we consider the family $\otimes_{[k;l]}$ which contracts the $k$th rank of the first tensor with the $l$th rank of the second tensor. This is only defined if the two ranks to be contracted have the same dimension, as follows: 
\begin{align*}
  &\left( \textbf{A} \otimes_{[k;l]} \textbf{B}\right)_{ \substack{i_1,\ldots ,i_{k-1}, j_1,\ldots ,j_{l-1},
            \\j_{l+1},\ldots ,j_m, i_{k+1},\ldots ,i_n}} \\ &\;\;\;:= \sum_{i_k, j_l} \delta(i_k, j_l) \textbf{A}_{i_1,\ldots ,i_n} \times \textbf{B}_{j_1,\ldots ,j_m},
\end{align*}
where $\delta$ is the identity function that is equal to $\1$ if $i_k = j_l$ and $\0$ otherwise. Note that the ranks corresponding to $\textbf{B}$ which are not contracted over go in between the ranks of $\textbf{A}$, replacing where the contracted rank of $\textbf{A}$ was. We will use $\otimes_{j}$ as a shorthand of $\otimes_{[j;1]}$, and in cases where $j=l=1$, we will omit the subscript on $\otimes$ altogether.

More generally, we will allow multiplication operations that contract multiple consecutive dimensions. $\mathbf{A} \otimes_{[k;l]}^r \mathbf{B}$ will denote contracting rank $k$ of $\mathbf{A}$ with rank $l$ of $\mathbf{B}$, rank $k+1$ of $\mathbf{A}$ with rank $l+1$ of $\mathbf{B}$ and so forth until rank $k+r-1$ of $\mathbf{A}$ and $l+r-1$ of $\mathbf{B}$. Formally:
\begin{align*}
&\left( \textbf{A} \otimes_{[k;l]}^r \textbf{B} \right)_{\substack{i_1,\ldots ,i_{k-1}, 
                    j_1,\ldots ,j_{l-1}, \\
                    j_{l+r},\ldots ,j_m, 
                    i_{k+r},\ldots ,i_n}} := \\& \sum_{\substack{i_{k}, \ldots , i_{k+r-1}\\
                                    j_{l},\ldots , j_{l+r-1}} }
                    \left( \prod_{p=0}^{r-1} \delta\left(i_{k+p}, j_{l+p}\right) \right) \textbf{A}_{i_1,\ldots ,i_n} \textbf{B}_{j_1,\ldots ,j_m}  
\end{align*}

We will use the notation $\mathbf{A} \otimes^* \mathbf{B}$ as a shorthand for $\mathbf{A} \otimes^{rank(\mathbf{A})} \mathbf{B}$ if $rank(\mathbf{A}) < rank(\mathbf{B})$ and $\mathbf{A} \otimes^{rank(\mathbf{B})} \mathbf{B}$ otherwise. 

To make the presentation clearer, we will also use the notation $X \otimes [A_1, A_2,\ldots ,A_k]$ to denote contraction of $A_1$ with the first rank of $X$, $A_2$ with the second and so forth. In other words $X \otimes [A_1,\ldots ,A_n]$ is equivalent to $X \otimes_n A_n \otimes_{n-1} A_{n-1} \ldots  \otimes_1 A_1$.

The multiplicative identity for $\mathbf{A} \in \mathbb{S}^{d_1 \times \ldots \times d_n}$ and $\otimes_{k}$ is the identity matrix $\mathbf{I} \in \mathbb{S}^{d_k \times d_k}$ where the diagonal entries are the multiplicative identity from the underlying semiring, and the non-diagonals are the additive identity. For $\mathbf{A} \in \mathbb{S}^{d_1 \times \ldots \times d_n}$ and $\otimes_{k}^r$ the multiplicative identity is a rank-$2r$ tensor $\mathbf{I} \in \mathbb{S}^{d_k \times \ldots \times d_{k+r-1} \times d_k \times \ldots \times d_{k+r-1}}$ and is defined as follows: 
\[\mathbf{I}_{d_1,\ldots ,d_r} = \prod_{i=0}^{\frac{n}{2}} \delta\left(d_i, d_{\frac{r}{2}+i}\right) \]

Lastly, as the higher order analogue of the transpose operator, we will define a permutation operator $\mathbf{A}^\pi$ where $\pi = [\pi_1, \pi_2, \ldots , \pi_r]$ is a permutation of $[1\ldots r]$ and $r$ is the rank of $\mathbf{A}$. The $\pi_i$th rank of $\mathbf{A}^\pi$ is equal to $i$th rank of $\mathbf{A}$. 

The key property of semirings for purposes of efficient calculation of item values is the distributive property. This property also holds for tensors over semirings.

\begin{lemma}
For any $k,l$, $\otimes_{[k;l]}$ distributes over $\oplus$
\end{lemma}

A proof can be found in Appendix A.

\subsection{Grammar Derivations}

For a grammar $G$ with a function $w$ that provides a mapping from rules to tensor weights, we will define a value of a derivation via the derivation tree: 

\begin{definition}
Given a grammar $G$ and a weight function $w$, the value of a derivation tree $T$ is:
\[ V_G^w(T) = 
    \begin{cases}
        w(r) \\
        \;\;\;\; \text{ if }  T =\langle r \rangle \\
        w(r) \otimes [V_G^w(T_1), \ldots , V_G^w(T_k)] \\ \;\;\;\; \text{ if } T = \langle r: T_1,\ldots ,T_k \rangle
    \end{cases} \]
\end{definition}

Note that there is no guarantee that this equation is defined for any arbitrary $w$. We will call a weight function $w$ \textit{well defined} for a grammar $G$ if for all valid derivation trees $T$ in $G$, $V_G^w(T)$ is defined. For CFGs there is a straightforward method to ensure that $w$ is well defined:

\begin{lemma} \label{lem:CFG-defined}
 A set of weights $w$ for a given CFG is well defined if there exist consistent dimensions $d_i$ for each nonterminal $A_i$ such that for all grammar rules $R: A_n \ra \alpha_1 A_1 \alpha_2 \ldots \alpha_{n-2} A_{n-1} \alpha_n$, $w(R) \in \mathbb{S}^{d_1 \times \ldots \times d_n} $
\end{lemma} 
Proof is given together with Lemma \ref{lem:DE-equiv}.

\begin{figure} 
\begin{small}
\begin{center}
    Tensor dimensions of grammar rules:
\end{center}
\begin{tabular}{l l} 
    $w(S \ra AA) \in \S^{A \times A \times S}$ 
    &$w(A \ra AA) \in \S^{A \times A \times A}$ \\ 
    $w(A \ra a) \in \S^A$ & 
\end{tabular}
\end{small}
\vspace{3mm} 

\begin{center} \small
    Grammar derivation tree:
\end{center}

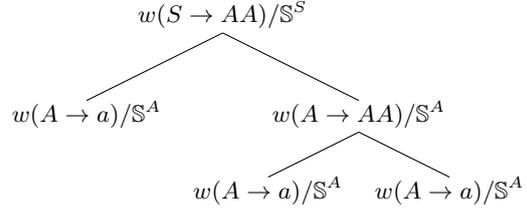
\parbox{1.5in}{\small \Tree 
    [.$w(S\ra AA)/\S^S$ 
        [.$w(A\ra a)/\S^A$ 
        ] 
        [.$w(A\ra AA)/\S^A$
            [.$w(A\ra a)/\S^A$ ] 
            [.$w(A\ra a)/\S^A$ ] 
        ] 
    ]
}

\begin{center}\small
    The value of the tree is given by the equation:
    \begin{align*} 
        w(S \ra &AA) \otimes (w(A \ra a), \\
                            &(w(A \ra AA) \otimes (w(A \ra a), w(A \ra a))))
    \end{align*}
\end{center}

\begin{center}\small
    Grammar derivation string:
\end{center}
\begin{align*}
    S \xRightarrow[\S^{A\times A \times S}]{ S\ra AA} AA
    \xRightarrow[\S^{A \times S}]{A \ra a} aA
    \xRightarrow[\S^{A\times A \times S}]{A\ra AA} aAA \\ 
    \xRightarrow[\S^{ A \times S}]{A\ra a} aaA
    \xRightarrow[\S^{S}]{ A\ra a} aaa
\end{align*}
\begin{center} \small
    The value of the string is given by the equation: 
    \begin{align*}
        w(S \ra AA) \otimes w(A \ra a) \otimes (A \ra AA) \\
        \otimes(A \ra a) \otimes (A \ra a)
    \end{align*}
\end{center}
\caption{\small Example derivation for the string ``aaa". We illustrate the initial dimensions of the tensor values for the rules and also show the intermediate tensor dimensions during the calculation of the value of the grammar tree and the grammar string.}

\vspace{-2ex}
\label{fig:grammar}
\end{figure}

Note that if a weight function for CFG is well defined, then the rank for the weights of rules with no non-terminals on their rhs is always 1. 

Given a grammar derivation tree $T$, 
let us call the list of derivation rules $E: R_1, R_2, \ldots , R_n$ appearing in $T$ ordered via depth-first, left-to-right manner a \textbf{grammar derivation string}.

\begin{definition} Given a CFG with tensor weights $w$, the
\textbf{value of a grammar derivation string} is defined as:
\[ V_G^w(E) = \bigotimes_{i} w(R_i) \]

\noindent where the application of $\otimes$ proceeds from left to right as is standard.
\end{definition}

For semirings, since the bracketing does not affect the final value of an expression, it is straightforward to show that the value of a grammar derivation tree corresponds to that of a grammar derivation string. With tensors over semirings this might fail with an arbitrary formalism $F$, and in the general we require the value of a derivation to be calculated with the bracketing induced by the derivation tree. However, for the special case of CFGs, the value of the grammar derivation tree and the value of its corresponding grammar derivation string are always equal. This means that for the computation of the value of the derivation, it is possible to replace the bracketing induced by the derivation tree by left-to-right bracketing without affecting the final value. Figure~\ref{fig:grammar} demonstrates the calculation of the value of the tree and the string for the same derivation together with how the tensor dimensions of the intermediate results evolve with each step of the calculation. 

\begin{lemma}\label{lem:DE-equiv}
Given a CFG $G$ and a weight function $w$ that fulfills the condition in Lemma \ref{lem:CFG-defined}, then $w$ is well defined and $V_G^w(T) = V_G^w(E)$ for any grammar derivation tree $T$ and corresponding grammar derivation string $E$.
\end{lemma}

\begin{proof}
We will proceed by induction on the derivation tree. If $T$ consists of only one rule $r$, then $V_G^w(T) = V_G^w(E)$. Furthermore, $r$ does not have any non-terminals on its rhs, so $V_G^w(T) \in \mathbb{S}^{d_0}$ with $ \mathbb{S}^{d_0}$ corresponding to the lhs non-terminal in $r$.

 Otherwise, $T$ has a labeled node $r$ and the subtrees $T_1 , \ldots , T_k$. Notice that if $A_0 \in \mathbb{S}^{d_1 \times \ldots \times d_n \times d_0}$, $A_1 \in \mathbb{S}^{d_2}$,\ldots , $A_n \in \mathbb{S}^{d_n}$, then $A_0 \otimes [A_1,\ldots ,A_n] = A_0 \otimes A_1 \otimes \ldots \otimes A_n$ due to all arguments within $[\ldots]$ being rank-1.
 
 Because $w$ fulfills the condition in Lemma \ref{lem:CFG-defined}, $w(r) \in \mathbb{S}^{d_1 \times \ldots \times d_k \times d_0}$ for some $d_i$ where $\mathbb{S}^{d_0}$ is the space corresponding to the non-terminal on the lhs of $r$, and $\mathbb{S}^{d_i}$ is the space corresponding to the $i$th non-terminal appearing in the rhs of $r$ for $i=1,\ldots ,k$. Then to complete the proof, it suffices to show that $V_G^w(T_i) \in \mathbb{S}^{d_i}$ for all subtrees $T_i$. This already holds for the base case.  For each $T_i: \langle r_i: T'_1,..,T'_k \rangle$, if $w(r_i) \in \mathbb{S}^{ d^i_1\times \ldots \times d^i_k \times d^i_0}$ then by induction $V_G^w(T_i) \in \mathbb{S}^{d^i_0}$, where $\mathbb{S}^{d^i_0}$ is the space corresponding to the non-terminal in the lhs of $R_i$. For the derivation to be valid, this non-terminal needs to match the $i$th non-terminal in the rhs of $R$, hence $\mathbb{S}^{d^i_0} = \mathbb{S}^{d_i}$
\end{proof}

\subsection{Item-based Descriptions}

Item-based descriptions are formal descriptions of various parsers for context-free grammars. Item-based descriptions consist of a set of deduction rules of the form $\displaystyle\frac{T_1\ldots T_k}{Q}P_1\ldots P_j$ where upper case letters could either be grammar rule templates (e.g. if $T_1:A \ra B\,\, C$ then any non-terminals from the grammar can be substituted for $A, B, C$) or for  \textbf{items}.
$T_1 \dots T_k$ are referred to as antecedents, $Q$ as the conclusion and $P_1 \ldots P_j$ are side conditions that the parser requires to execute the rule, but doesn't use the values of. Items correspond to chart elements in procedural descriptions of parsers, and are placeholders for intermediate results which can be combined to obtain the final result. The item-based description also provides a special \textbf{goal item} which is variable-free, and does not occur as a condition of any other inference rules.

\begin{figure*}
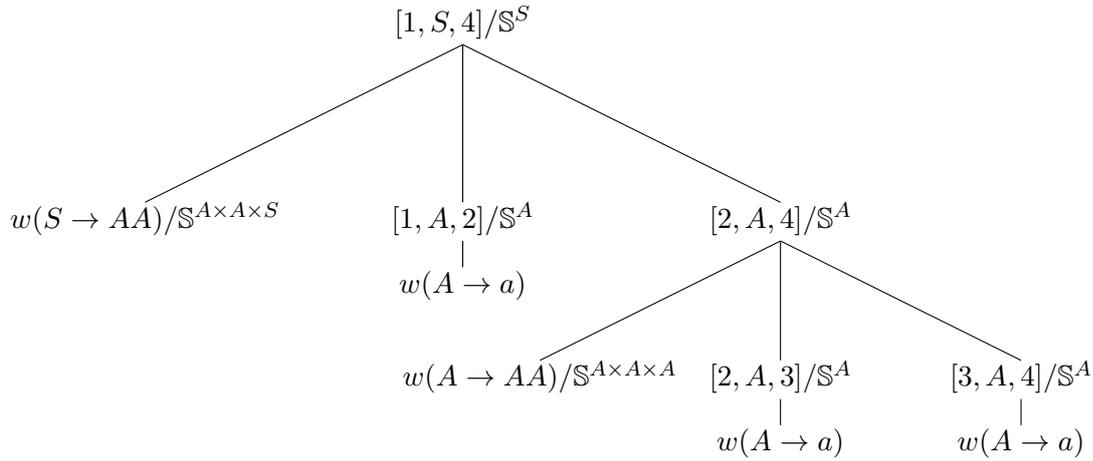

    \parbox{1.5in}{ \Tree 
    [.$[1,S,4]/\S^S$ 
        [.$w(S\ra AA)/\S^{A\times A\times S}$ ] 
        [.$[1,A,2]/\S^A$ $w(A\ra a)$ ] !\qsetw{-2cm} 
        [.$[2,A,4]/\S^A$ 
            [.$w(A\ra AA)/\S^{A\times A\times A}$ ] 
            [.$[2,A,3]/\S^A$ $w(A\ra a)$ ] 
            [.$[3,A,4]/\S^A$ $w(A\ra a)$ ]
        ]          
    ]
}

\caption{\small Item derivation corresponding to the derivation given in Figure \ref{fig:grammar} using the item-based description of CKY in Figure \ref{fig:CKY}.} 
\vspace{-2ex}
\label{fig:item-deriv}
\end{figure*}

\begin{figure} 
    \begin{small}
        \begin{equation*}
        \dfrac{w(A \ra w_i)}{[i, A, j]} 
    \end{equation*}
    \begin{equation*}
        \dfrac{w(A \ra B C) \hspace{2mm} [i, B, k] \hspace{2mm} [k,C,j]}{[i,A,j]}
    \end{equation*}
    \caption{\small Item-based description for CKY} \label{fig:CKY}
    \end{small}
\end{figure}

\begin{definition}
Given a grammar $G$ and an item-based description $I$, a valid \textbf{item derivation tree} is defined as follows:
\begin{itemize}
    \item For all $r \in G$, $\langle r \rangle$ is an item derivation tree.
    \item If $D_{a_1},\ldots ,D_{a_k}$ and $D_{c_1},\ldots ,D_{c_j}$ are derivation trees headed by $a_1,\ldots ,a_k$ and $c_1,\ldots ,c_j$ respectively, and $\frac{a_1\ldots a_k}{b}c_1,\ldots ,c_j$ is the instantiation of a deduction rule in $I$, then $\langle b: D_{a_1},\ldots ,D_{a_k} \rangle$ is also an item derivation tree.
\end{itemize}
\end{definition}

$inner_\sigma(x)$ denotes the set of all trees headed by $x$ that occur in parses for $\sigma$. Formally, $D \in inner_\sigma(x)$ if $D$ is headed by $x$ and is a subtree of some $D' \in \mathcal{D}_{I(G)}(\sigma)$. The value of a derivation tree is calculated similarly to that of a grammar tree: 
\begin{align*}
    &V_{I(G)}^w(D) = \\
       &\begin{cases} 
            w(D) \hspace{1.1 cm}\text{ if $D$ is a rule} \\
            V_{I(G)}^w(D_1) \otimes [V_{I(G)}^w(D_2), \ldots, V_{I(G)}^w(D_n)] \\ \hspace{2 cm}\text{ if }D = \langle b: D_1, \ldots, D_n \rangle
       \end{cases}
\end{align*}

Notice that unlike the definition from \citet{Goodman1999SemiringParsing}, the first antecedent in the inference rule has a special role in the calculation. Intuitively, our framework treats the value of the first antecedent as a \textit{function}, and the trailing ones as the arguments. The interaction between the trailing antecedents is thus moderated 
through the value of the first antecedent, which corresponds to the requirement that the children nodes be independent of each other 
given the parent node. 
 
\begin{definition}
If for any $\sigma \in \mathcal{L}(G)$ and any $T, T' \in inner_\sigma(x)$, $V^w_{I(G)}(T)$ and $V^w_{I(G)}(T')$ are defined and $dim(V^w_{I(G)}(T)) = dim(V^w_{I(G)}(T'))$, then the weights $w$ are well defined. 
\end{definition}

Given an item-based derivation $I$, a grammar $G$,
a well defined weight function $w$ and a target sentence $\sigma$, the value of an item $x$ is defined to be the sum of all its possible derivations. Formally:
\[V_{I(G)}^w(x, \sigma) = \bigoplus_{D \in inner_\sigma(x)} V_{I(G)}^w(D) \]

\begin{definition}
For a given grammar $G$ and item-based description $I$, the value of a sentence $\sigma$ is equal to the value of the goal item which spans $\sigma$:
\[ V_{I(G)}^w(\sigma) = V_{I(G)}^w(goal, \sigma) \]
\end{definition}

\begin{definition}
An item-based description is \textbf{correct} if for all grammars $G$, complete semirings $\S$, well defined weight functions $w$ and sentences $\sigma$,
$V_{I(G)}^w(\sigma) = V_G^w(\sigma)$
\end{definition}

Now we are ready to state the equivalent theorem to Theorem \ref{thm:correctness-goodman}. Let us introduce a special symbol $\bot$ and extend $V^w_G$ and $V^w_{I(G)}$ to any weight function $w$ so that if $w$ is not-well defined for $G$, then $V^w_G(\sigma) = \bot$ and likewise for $V^w_{I(G)}$. 

\begin{theorem}
 An item-based description $I$ is correct if
 \begin{itemize}
     \item For every grammar $G$, the mapping $g: \mathcal{D}_{I(G)} \ra \mathcal{D}_G$ that maps $d' \in \mathcal{D}_{I(G)}$ to the corresponding $d \in \mathcal{D}_G$ is a bijection with an inverse function $f$.
     
     \item For any complete semiring $\S$ and weight function $w$, $g$ and $f$ preserve the values assigned to a derivation: 
     \begin{align*}
         V_G^w(d) &= V_{I(G)}^w(f(d)) \text{ and } \\ V_{I(G)}^w(d') &= V_G^w(g(d'))
     \end{align*}
 \end{itemize}
\end{theorem}

Proof proceeds similarly to that in \cite{Goodman1999SemiringParsing} and can be found in Appendix A.

\section{Inside and Outside Calculations}
In the following, we will omit the sentence $\sigma$ from $inner_\sigma(x)$ and refer to this as $inner(x)$. Let $inner(\frac{a_1,..,a_k}{x})$ the set of derivation trees where the root note is $x$, and the direct children of $x$ are $a_1,\ldots ,a_k$.

For efficient computation of this value, we will assume that there is a partial order $b$ on the items so that if the item $y$ depends on $x$, then $b(x) \leq b(y)$. 

\begin{theorem}
\begin{align*}
    V(x) = \bigoplus_{\substack{[a_1,\ldots ,a_k] \\ \text{s.t.}\frac{a_1,..,a_k}{x}}} &V(a_1) \otimes \left[  V(a_2) , \ldots ,V(a_k) \right]
\end{align*}
\end{theorem}

The proof uses the distributive property and follows that of \citet{Goodman1999SemiringParsing}. It can be found in Appendix A.

For the notion of a value of a derivation to extend to outside trees, we will have to do some modifications. This is because an outside tree will have one subtree $\langle b: A_1,\ldots ,A_n \rangle$,  such that $V(A_1) \otimes [V(A_2), \ldots , V(A_n)]$ will potentially not be defined since one of the subtrees $A_k$ will be missing. Note that the missing $A_k$ will be headed by an item. We will say the a tree $T \in outer(x)$ if $T$ can be obtained by taking a tree $T'$ headed by the goal item and removing any of its subtrees headed by the item $x$. Outer value $Z(T_k)$ is defined recursively as follows:

If $T_k$ is headed by the goal item then $Z(T_k) = I_{d_S}$. Else, it has a direct parent tree $T$ such that $T = \langle b: T_1,\ldots ,T_k,\ldots ,T_n \rangle$. In this case, $Z(T_k) = $
\vspace{-1mm}
\begin{align*}
    &\Big( V(T_1) 
     \otimes_k \left[ I_{T_k \times d_S}, V(T_{k+1}),\ldots ,V(T_n) \right] \Big)^\pi \\
    &\otimes \left[ V(T_2),\ldots ,V(T_{k-1}) \right] \otimes^* Z(T)
\end{align*}
where $I_{T_k \times d_S}$ is the identity tensor for the  space $\mathbb{S}^{d_1 \times \ldots \times d_i \times d_S}$, $T_k \in \mathbb{S}^{d_1 \times \ldots \times d_i}$, and $d_s$ is the dimension assigned for the terminal symbol $S$. 
The permutation $\pi$ is defined as follows: 
\[[1,2,\ldots ,i, j+1, j+2,\ldots ,n, i+1, i+2,\ldots ,j]\] 
where $i = k + \text{rank}(T_k) - 1$ and $j = k + 2 \times \text{rank}(T_k) +1$

To understand the function of $\pi$ it is useful to consider the dimensions of the term before and after it is applied. Let the term $V(T_0) \otimes_k \left[ I_{T_k \times d_S}, V(T_{k+1}),\ldots ,V(T_n) \right]$ have dimensions:
\begin{align*}
    &e_1 \times \ldots \times e_{k-1}, d_1 \times \ldots \times d_i \times d_S \times \\ &e_k \times d_1 \times \ldots \times d_i\times d_S \times d'_n \times \ldots \times d'_m 
\end{align*}
Here $e_1,\ldots , e_{k-1}$ are the dimensions that will be contracted with $V(T_1),\ldots , V(T_{k-1})$ with the second multiplication operation, and $d'_n, \ldots , d'_m$ are the dimensions that were either introduced by the contraction with $V(T_{k+1}),\ldots ,V(T_n)$ or were trailing dimensions from $V(T_1)$. The result of the contraction with $I_{T_k \times d_S}$ are the dimensions in the middle: $d_1,\ldots ,d_i, d_S, e_k, d_1,\ldots , d_i, d_S$. Unlike the original definition of $I$ there is one dimension $e_k$ missing from the beginning of the sequence since it got used up during the contraction operation. What the permutation does is to move one section of the dimensions introduced by $I$ to the very end. The dimensions become:
\begin{align*}
    &e_1 \times \ldots \times e_{k-1}, d_1 \times\ldots  \times d_i \times \\ & d'_n \times \ldots \times d'_m \times d_S \times e_k \times d_1 \times \ldots \times d_i \times d_S 
\end{align*}
Note that this has no effect on the next contraction with $V(T_1),\ldots , V(T_{k-1})$ since the first $k-1$ ranks are left in place. However, changing the order of the ranks allow the last contraction with $Z(T)$ to be well defined. 

\begin{lemma}\label{lem:inside-outside-tree} Let $V$ and $Z$ be defined on a \textit{commutative} semiring $\mathbb{S}$ and
let $O \in outer_\sigma(x)$ and $T \in inner_\sigma(x)$. If combining $O$ and $T$ in the obvious way results in the complete derivation $D$,  
\[V(D) = V(T) \otimes^* Z(O) \]
\end{lemma}
\begin{proof} (Sketch) 
We proceed by induction on the parse tree. Base case is where $x = goal$, $T = D$ and $O$ is empty. Then $V(T) = V(D)$ and $Z(O) = I_S$. $V(D) \otimes^* I_S = V(D)$ by the definition of $I_S$ which proves the statement. 

Otherwise $T$ has a parent tree $T_p = \langle y: T_1, \ldots , T_n \rangle$ where $T = T_k$. Furthermore, $T_p \in \text{inner}_\sigma(y)$, $O_p \in \text{outer}_\sigma(y)$ and by
the induction hypothesis $V(D) = V(T_p) \otimes^* Z(O_p)$.

Since $T_p \in \text{inner}_\sigma(y)$ we know that 
\begin{equation*}
    V(T_p) = V(T_1) \otimes \left[V(T_2), \ldots , V(T_m) \right] 
\end{equation*} $V(D) =$
\begin{equation*}
    (V(T_1) \otimes [V(T_2), \ldots , V(T_m)]) \otimes^* Z(O_p)
\end{equation*}
The proof progresses by calculating the value for $[ V(D)]_i$ based on the above term and shows that this is equal to the value of $[V(T) \otimes^* Z(O)]_i$. Full proof can be found in Appendix A.
\end{proof}

In the general case, \citet{Goodman1999SemiringParsing} defines the reverse value of $x$ as the sum of all its outer trees.
\[Z(x) = \bigoplus_{T \in outer(x)} Z(T) \]


We will see that for a well defined weight function $w$,
any $D \in outer_\sigma(x)$ will be assigned a value with dimensions $d_1 \times \ldots \times d_n \times d_S$ where $d_S$ is the dimension assigned to the start symbol $S$, and $d_1,\ldots ,d_n$ are the dimensions for $inner_\sigma(x)$.

\begin{lemma}
Let $C(D, x)$ represent the number of times $x$ occurs in a derivation $D$. Then,
\[V(x) \otimes^{*} Z(x) = \bigoplus_{D \in \mathcal{D}(\sigma)} V(D)C(D,x)\]
\end{lemma}

\begin{proof}
\begin{align*}
    &V(x) \otimes^* Z(x) =  \hspace{-2mm} \bigoplus_{T \in inner(x)} \hspace{-2mm} V(T)  \otimes^* \hspace{-2mm} \bigoplus_{O \in outer(x)} \hspace{-2mm} Z(O)   \\
        &= \bigoplus_{T \in inner(x)} \bigoplus_{O \in outer(x)} V(T)\otimes^* Z(O) 
\end{align*}
By Lemma \ref{lem:inside-outside-tree}, $Z(O) \otimes^* V(T) = V(D)$. For an item $x$, any $O \in outer(x)$ and $T \in inner(x)$ can be combined to form a successful derivation tree containing $x$, and thus the number $C(D, x)$ corresponds exactly to the number of derivation trees containing $x$. Hence,
\begin{align*}
    V(x) \otimes^* Z(X) &= \hspace{-2mm}\bigoplus_{\substack{T \in inner(x) \\ O \in outer(x)}}  V(T)\otimes^* Z(O) \\ &=  \bigoplus_{D \in \mathcal{D}(\sigma)} V(D)C(D,x) \qedhere
\end{align*}
\end{proof}

Now we are ready to state how to calculate the outside value of an item. Following \citet{Goodman1999SemiringParsing} we will extend the notation for the set of outer trees and introduce $outer \left(k, \frac{a_1\ldots a_n}{b} \right) \subseteq outer(a_k)$ to mean the subset of the outer trees in $outer(a_k)$ where $a_k$ has parent $b$ and the siblings $a_i$. In other words, this is the set of all outer trees where the rule from which $a_k$ is removed is $\displaystyle\frac{a_1\ldots a_n}{b}$. 

\begin{theorem}
  If $x$ is the goal item, then $Z(x) = I_s$. Else, $Z(x) =$ 
 \begin{align*}
     &\bigoplus_{\substack{j,a_1,..,a_k,b \text{ s.t. } \\ \frac{a_1\ldots a_k}{b} \text{ and } x=a_j}} \hspace{-6mm} (V(a_1) \otimes_k \left[I_{a_k}, V(a_{k+1}), ... ,V(a_n) \right])^\pi  \\ &\hspace{2.4cm} \otimes \left[V(a_2),\ldots ,V(a_{k-1}) \right] \otimes^* Z(b) 
 \end{align*}
\end{theorem}
\begin{proof}(sketch) $Z(x) = \bigoplus_{D \in outer(x)} Z(D)$. Either $x$ is a goal item, in which case $Z(x) = I_S$. 

Otherwise the outer trees $outer(x)$ could be written as the union of outer trees $outer\left(k, \frac{a_1\ldots a_n}{b} \right)$ for each rule $\frac{a_1\ldots a_n}{b}$ where $a_k = x$ for some $k$. Hence:
\[Z(x) = \bigoplus_{\substack{j,a_1,..,a_k,b \text{ s.t. } \\ \frac{a_1\ldots a_k}{b} \text{ and } x=a_j}} \bigoplus_{D \in outer\left(k,\frac{a_1\ldots a_n}{b}\right)} Z(D) \]
Using the distributive property of the partial semiring, the inside part of the equation becomes:
\begin{align*}
    &\bigoplus_{D \in outer\left(k,\frac{a_1\ldots a_n}{b}\right)} Z(D) = \\  &\hspace{1cm} \left(V(a_1) \otimes_k \left[I_{a_k}, V(a_{k+1}),\ldots ,V(a_n) \right]\right)^\pi  \\ &\hspace{1cm} \otimes  \left[V(a_2),\ldots ,V(a_{k-1}) \right] \otimes^* Z(b) 
\end{align*}

Replacing the inner part of the previous equation with this term gives the desired equality.
\end{proof}

\section{Conclusion}
We have presented a general extension of the semiring parsing framework where the weights for the grammar rules are tensors of semiring values, with the motivation of extending semiring parsing framework to latent variable models. We hope that this work will enable streamlined development of EM-based or spectral learning algorithms for latent refinements of a number of grammar formalisms.

\section*{Acknowledgments}

The authors thank 
the anonymous reviewers for feedback and comments on a draft of this paper,
and acknowledge the support of NSF grant IIS-1813823.

\bibliography{references1}

\begin{thebibliography}{22}
\expandafter\ifx\csname natexlab\endcsname\relax\def\natexlab#1{#1}\fi

\bibitem[{Bailly et~al.(2009)Bailly, Denis, and
  Ralaivola}]{Bailly2009GrammaticalProblem}
Raphaël Bailly, François Denis, and Liva Ralaivola. 2009.
\newblock {Grammatical inference as a principal component analysis problem}.
\newblock In \emph{Proceedings of the 26th Annual International Conference on
  Machine Learning}, pages 33--40.

\bibitem[{Boullier(2004)}]{Boullier2004RangeGrammars}
Pierre Boullier. 2004.
\newblock {Range concatenation grammars}.
\newblock In \emph{New Developments in Parsing Technology}, pages 269--289.
  Springer.

\bibitem[{Cohen et~al.(2008)Cohen, Simmons, and
  Smith}]{Cohen2008DynamicPrograms}
Shay~B Cohen, Robert~J Simmons, and Noah~A Smith. 2008.
\newblock {Dynamic programming algorithms as products of weighted logic
  programs}.
\newblock In \emph{International Conference on Logic Programming}, pages
  114--129.

\bibitem[{Cohen et~al.(2013)Cohen, Stratos, Collins, Foster, and
  Ungar}]{Cohen2013ExperimentsPCFGs}
Shay~B Cohen, Karl Stratos, Michael Collins, Dean~P Foster, and Lyle Ungar.
  2013.
\newblock \href {https://www.aclweb.org/anthology/N13-1015} {{Experiments with
  Spectral Learning of Latent-Variable PCFGs}}.
\newblock In \emph{Proceedings of the 2013 Conference of the North American
  Chapter of the Association for Computational Linguistics: Human Language
  Technologies}, pages 148--157, Atlanta, Georgia. Association for
  Computational Linguistics.

\bibitem[{Cohen et~al.(2014)Cohen, Stratos, Collins, Foster, and
  Ungar}]{Cohen2014SpectralComplexity}
Shay~B Cohen, Karl Stratos, Michael Collins, Dean~P Foster, and Lyle Ungar.
  2014.
\newblock {Spectral learning of latent-variable PCFGs: Algorithms and sample
  complexity}.
\newblock \emph{The Journal of Machine Learning Research}, 15(1):2399--2449.

\bibitem[{Dempster et~al.(1977)Dempster, Laird, and
  Rubin}]{Dempster1977MaximumAlgorithm}
Arthur~P Dempster, Nan~M Laird, and Donald~B Rubin. 1977.
\newblock {Maximum likelihood from incomplete data via the EM algorithm}.
\newblock \emph{Journal of the Royal Statistical Society: Series B
  (Methodological)}, 39(1):1--22.

\bibitem[{Eisner(2002)}]{Eisner2002ParameterTransducers}
Jason Eisner. 2002.
\newblock \href {https://doi.org/10.3115/1073083.1073085} {{Parameter
  Estimation for Probabilistic Finite-State Transducers}}.
\newblock In \emph{Proceedings of the 40th Annual Meeting of the Association
  for Computational Linguistics}, pages 1--8, Philadelphia, Pennsylvania, USA.
  Association for Computational Linguistics.

\bibitem[{Eisner et~al.(2005)Eisner, Goldlust, and
  Smith}]{Eisner2005CompilingLanguage}
Jason Eisner, Eric Goldlust, and Noah~A Smith. 2005.
\newblock \href {https://www.aclweb.org/anthology/H05-1036} {{Compiling Comp
  Ling: Weighted Dynamic Programming and the Dyna Language}}.
\newblock In \emph{Proceedings of Human Language Technology Conference and
  Conference on Empirical Methods in Natural Language Processing}, pages
  281--290, Vancouver, British Columbia, Canada. Association for Computational
  Linguistics.

\bibitem[{Gebhardt(2018)}]{Gebhardt2018GenericParsing}
Kilian Gebhardt. 2018.
\newblock \href {https://www.aclweb.org/anthology/C18-1258} {{Generic
  refinement of expressive grammar formalisms with an application to
  discontinuous constituent parsing}}.
\newblock In \emph{Proceedings of the 27th International Conference on
  Computational Linguistics}, pages 3049--3063, Santa Fe, New Mexico, USA.
  Association for Computational Linguistics.

\bibitem[{Gimpel and Smith(2009)}]{Gimpel2009CubeSemirings}
Kevin Gimpel and Noah~A Smith. 2009.
\newblock \href {https://www.aclweb.org/anthology/E09-1037} {{Cube Summing,
  Approximate Inference with Non-Local Features, and Dynamic Programming
  without Semirings}}.
\newblock In \emph{Proceedings of the 12th Conference of the European Chapter
  of the ACL (EACL 2009)}, pages 318--326, Athens, Greece. Association for
  Computational Linguistics.

\bibitem[{Goodman(1999)}]{Goodman1999SemiringParsing}
Joshua Goodman. 1999.
\newblock \href {https://www.aclweb.org/anthology/J99-4004} {{Semiring
  Parsing}}.
\newblock \emph{Computational Linguistics}, 25(4):573--606.

\bibitem[{Goodman(1998)}]{Goodman1998ParsingInside-Out}
Joshua~T Goodman. 1998.
\newblock \emph{{Parsing Inside-Out}}.
\newblock Ph.D. thesis, Harvard University Cambridge, Massachusetts.

\bibitem[{Hsu et~al.(2012)Hsu, Kakade, and Zhang}]{Hsu2012AModels}
Daniel Hsu, Sham~M Kakade, and Tong Zhang. 2012.
\newblock {A spectral algorithm for learning hidden Markov models}.
\newblock \emph{Journal of Computer and System Sciences}, 78(5):1460--1480.

\bibitem[{Kuich(1997)}]{Kuich1997SemiringsAutomata}
Werner Kuich. 1997.
\newblock \href {https://doi.org/10.1007/978-3-642-59136-5{\_}9} {{Semirings
  and Formal Power Series: Their Relevance to Formal Languages and Automata}}.
\newblock In Rozenberg Grzegorz and Arto Salomaa, editors, \emph{Handbook of
  Formal Languages: Volume 1 Word, Language, Grammar}, pages 609--677.
  Springer, Berlin, Heidelberg.

\bibitem[{Li and Eisner(2009)}]{Li2009First-Forests}
Zhifei Li and Jason Eisner. 2009.
\newblock \href {https://www.aclweb.org/anthology/D09-1005} {{First- and
  Second-Order Expectation Semirings with Applications to Minimum-Risk Training
  on Translation Forests}}.
\newblock In \emph{Proceedings of the 2009 Conference on Empirical Methods in
  Natural Language Processing}, pages 40--51, Singapore. Association for
  Computational Linguistics.

\bibitem[{Lopez(2009)}]{Lopez2009TranslationDeduction}
Adam Lopez. 2009.
\newblock \href {https://www.aclweb.org/anthology/E09-1061} {{Translation as
  Weighted Deduction}}.
\newblock In \emph{Proceedings of the 12th Conference of the European Chapter
  of the ACL (EACL 2009)}, pages 532--540, Athens, Greece. Association for
  Computational Linguistics.

\bibitem[{Matsuzaki et~al.(2005)Matsuzaki, Miyao, and
  Tsujii}]{Matsuzaki2005ProbabilisticAnnotations}
Takuya Matsuzaki, Yusuke Miyao, and Jun'ichi Tsujii. 2005.
\newblock {Probabilistic CFG with latent annotations}.
\newblock In \emph{Proceedings of the 43rd Annual Meeting of the Association
  for Computational Linguistics}, pages 75--82.

\bibitem[{Nederhof(2003)}]{Nederhof2003WeightedAlgorithm}
Mark-Jan Nederhof. 2003.
\newblock \href {https://doi.org/10.1162/089120103321337467} {{Weighted
  Deductive Parsing and Knuth's Algorithm}}.
\newblock \emph{Computational Linguistics}, 29(1):135--143.

\bibitem[{Pereira and Warren(1983)}]{Pereira1983ParsingDeduction}
Fernando C~N Pereira and David H~D Warren. 1983.
\newblock \href {https://doi.org/10.3115/981311.981338} {{Parsing as
  Deduction}}.
\newblock In \emph{Proceedings of the 21st Annual Meeting on Association for
  Computational Linguistics}, pages 137--144, Cambridge, Massachusetts, USA.
  Association for Computational Linguistics.

\bibitem[{Shieber et~al.(1995)Shieber, Schabes, and
  Pereira}]{Shieber1995PrinciplesParsing}
Stuart~M Shieber, Yves Schabes, and Fernando C~N Pereira. 1995.
\newblock {Principles and implementation of deductive parsing}.
\newblock \emph{The Journal of logic programming}, 24(1-2):3--36.

\bibitem[{Sikkel(1998)}]{Sikkel1998ParsingAlgorithms}
Klaas Sikkel. 1998.
\newblock {Parsing schemata and correctness of parsing algorithms}.
\newblock \emph{Theoretical Computer Science}, 199(1-2):87--103.

\bibitem[{Steenstrup(1985)}]{Steenstrup1985Sum-OrderedSemirings}
Martha~Edmay Steenstrup. 1985.
\newblock \emph{{Sum-Ordered Partial Semirings}}.
\newblock Ph.D. thesis, University of Massachusetts Amherst.

\end{thebibliography}

\bibliographystyle{acl_natbib}

\newpage
\onecolumn

\section*{Appendix A - Proofs of Theorems in Main Paper}

\begin{manuallemma}{5.1}
For any $k,l$, $\otimes_{[k;l]}$ distributes over $\oplus$
\end{manuallemma}

\begin{proof}
We will proceed by showing that:
\[A \otimes_{[k;l]}(B \oplus C) = (A \otimes_{[k;l]} B) \oplus (A \otimes_{[k;l]} C)\]

Firstly, note that for the left hand side of the equation to be defined, $B$ and $C$ needs to be of matching ranks, and that $B \oplus C$ will be the same rank as both $B$ and $C$. Therefore, if the left hand side is well defined then both $A \otimes_{[k;l]} B$ and $A \otimes_{[k;l]} C$ is defined and has matching ranks. So the right hand side is defined if and only if the left hand side is defined as well.

\begin{align*}
    [A \otimes_{[j;k]} &(B \oplus C)]_{i_1,\ldots ,i_{k-1}, j_1,\ldots ,j_{l-1},j_{l+1},\ldots ,j_m, i_{k+1},\ldots ,i_n} \\
    &= \sum_{i_k,j_l} \delta(i_k,j_l) A_{i_1,\ldots ,i_n} \times (B \oplus C)_{j_1,\ldots ,j_m}\\ 
    &= \sum_{i_k,j_l} \delta(i_k,j_l) A_{i_1,\ldots ,i_n} \times (B_{j_1,\ldots ,j_m} + C_{j_1,\ldots ,j_m})\\
    &= \sum_{i_k,j_l} \delta(i_k,j_l) (A_{i_1,\ldots ,i_n} \times B_{j_1,\ldots ,j_m})     + \delta(i_k,j_l) (A_{i_1,\ldots ,i_n} \times C_{j_1,\ldots ,j_m})\\
    &= [(A \otimes_{[k;l]} B) \oplus (A \otimes_{[k;l]} C)]_{i_1,\ldots ,i_{k-1}, j_1,\ldots ,j_{l-1},j_{l+1},\ldots ,j_m, i_{k+1},\ldots ,i_n}
\end{align*}
\end{proof}

\begin{manualtheorem}{5.4}
 An item-based description $I$ is correct if
 \begin{itemize}
      \item For every grammar $G$, the mapping $g: \mathcal{D}_{I(G)} \ra \mathcal{D}_G$ that maps $d' \in \mathcal{D}_{I(G)}$ to the corresponding $d \in \mathcal{D}_G$ is a bijection with an inverse function $f$.
     
     \item For any complete semiring $S$ and weight function $w$, $g$ and $f$ preserve the values assigned to a derivation: 
     \[V_G^w(d) = V_{I(G)}^w(f(d)) \text{ and } V_{I(G)}^w(d') = V_G^w(g(d'))\] 
 \end{itemize}
\end{manualtheorem}

\begin{proof}
\begin{align*}
    V_{I(G)}^w(\alpha) &= V_{I(G)}^w(goal, \alpha) = \bigoplus_{D \in \text{inner}_\alpha(goal)} V_{I(G)}^w(D) 
    = \bigoplus_{D \in \mathcal{D}_{I(G)}(\alpha)} V_G^w(g(D))
\end{align*}
Observe that $D \in \mathcal{D}_{I(G)}(\alpha) $ iff $g(D) \in \mathcal{D}_G(\alpha)$ since the rules that appear in the leaves of $D$, applied from left to right, determines the grammar derivation tree $g(D)$ uniquely via $g$, and vice versa. Hence,
\begin{align*}
    V_{I(G)}^w(\alpha) &= \bigoplus_{g(D) \in \mathcal{D}_{G}(\alpha)} V_G^w(g(D)) = V_{G}^w(\alpha)
\end{align*}
\end{proof}

\begin{manualtheorem}{6.1}
\begin{align*}
    V(x) = \bigoplus_{\substack{[a_1,\ldots ,a_k] \\ \text{s.t.}\frac{a_1,..,a_k}{x}}} &V(a_1) \otimes \left[  V(a_2) , \ldots ,V(a_k) \right]
\end{align*}
\end{manualtheorem}
\begin{proof}
Recall that by definition, $V(x) = \bigoplus_{D \in inner(x)} V(D)$.
For any item derivation $D$, $D$ is either an axiom or there is some $a_1,\ldots ,a_k,b$ s.t. $D \in inner(\frac{a_1 \ldots  a_k}{b})$. If $D$ is an axiom, 
then $inner(D)$ is just a single rule $a$, and so 
$V(D) = V(a)$. Else, for each rule $\frac{a_1\ldots a_k}{x}$
\begin{align*}
&\bigoplus_{D \in inner(\frac{a_1\ldots a_k}{x})} V(D) =            
    \bigoplus_{\substack{D_{a_1} \in inner(a_1),\ldots , \\
                         D_{a_k}\in inner(a_k)}}  \hspace{-3mm}
    V(D_{a_1}) \otimes [  V(D_{a_2}),\ldots  ,V(D_{a_k})]  \\
    &= \left( \bigoplus_{D_{a_1} \in inner(a_1)}  V(D_{a_1}) \right) \otimes
             \left( \bigoplus_{\substack{D_{a_2}\in inner(a_2), \ldots, \\ D_{a_k} \in inner(a_k)}}
            \bigotimes_{i=2}^k V(D_{a_i}) \right) \\
    &= \left( \bigoplus_{D_{a_1} \in inner(a_1)}  V(D_{a_1}) \right) \otimes
        \left( \bigoplus_{D_{a_2} \in inner(a_2)} V(D_{a_2}), \ldots  , \bigoplus_{D_{a_k} \in inner(a_k)} V(D_{a_k})\right) \\
    &= V(a_1) \otimes \left[V(a_2),\ldots , V(a_k)\right]
\end{align*}
Where the last step holds due to the distributive property of the partial semiring. 

Since the set $inner(x) = \bigcup_i D_i$ where $D_i \in inner(\frac{a_1\ldots a_k}{x})$ for all inference rules $\frac{a_1\ldots a_k}{x}$, we can write the summation over $D \in inner(x)$ as:
\begin{align*}
    &V(x) = \bigoplus_{D \in inner(x)} V(D) \\
         &= \bigoplus_{\substack{[a_1,\ldots ,a_k] \\ \text{s.t.}\frac{a_1,..,a_k}{x}}} 
            \bigoplus_{D \in inner(\frac{a_1\ldots a_k}{x})} V(D) \\
        &= \bigoplus_{\substack{[a_1,\ldots ,a_k] \\ \text{s.t.}\frac{a_1,..,a_k}{x}}} V(a_1) \otimes \left[  V(a_2), V(a_3),\ldots ,V(a_k) \right]
\end{align*}
Where the last line is obtained by replacing the inner part of the expression with the equality obtained from the previous part of the proof.
\end{proof}

\begin{manuallemma}{6.2}
Let $V$ and $Z$ be defined on a \textit{commutative} semiring $\mathbb{S}$ and
let $O \in outer_\alpha(x)$ and $T \in inner_\alpha(x)$. If combining $O$ and $T$ in the obvious way results in the complete derivation $D$ then  
\[V(D) = V(T) \otimes^* Z(O) \]
\end{manuallemma} 

\begin{proof}
To simplify notation of the indices, let $\mathbf{i}$ stand for a list of indices $i_1,\ldots ,i_n$ for some $n$. We will also use $\dd^i$ to denote a list $d_1^i,\ldots d_{n_i}^i$ and $\dd$ to denote $\dd^1,\ldots ,\dd^n$. $\delta(\mathbf{i}, \mathbf{j}) = \prod_{k=1}^n \delta(i_k, j_k)$.

We will proceed by induction on the parse tree. Base case is where $x = goal$, $T = D$ and $O$ is empty. Then $V(T) = V(D)$ and $Z(O) = I_S$. $V(D) \otimes^* I_S = V(D)$ by the definition of $I_S$ which proves the statement. 

Otherwise $T$ has a parent tree $T_p = \langle y: T_1, \ldots , T_n \rangle$ where $T = T_k$. Furthermore, $T_p \in \text{inner}_\alpha(y)$, $O_p \in \text{outer}_\alpha(y)$ and by induction hypothesis $V(D) = V(T_p) \otimes^* Z(O_p)$.

Since $T_p \in \text{inner}_\alpha(y)$ we know that 
\[V(T_p) = V(T_1) \otimes \left[V(T_2), \ldots , V(T_m) \right] \] 
So
\[V(D) = \left(V(T_1) \otimes [V(T_2), \ldots , V(T_m)] \right) \otimes^* Z(O_p) \]

The proof progresses by calculating the value for $[ V(D)]_i$ based on the above term and shows that this is equal to the value of $[V(T) \otimes^* Z(O)]_i$.

Let:
\begin{align*}
    &V(T_1) \in \mathbb{S}^{\ee, \ff}
    &V(T_i) \in \mathbb{S}^{e_i, \dd^i} \\
    &Z(O_p) \in \mathbb{S}^{\dd, \ff, s} 
    &V(D) \in \mathbb{S}^s
\end{align*}
Then:
\begin{align*}
    V[(T_p)]_{\dd,\ff} &= 
    \left[V(T_1) \otimes \left(V(T_2),\ldots ,V(T_m) \right) \right]_{\dd,\ff}\\ &= 
    \sum_{\ee, \ee'} V(T_1)_{\ee, \ff} \times \prod_{i=2}^m \delta(e_i, e_i') V(T_i)_{e_i', \dd^i}
\end{align*}
\begin{align*}
    [V(D)]_s = \left[V(T_p) \otimes^* Z(O_p)\right]_s = &\\ 
    \sum_{\ee, \mathbf{e'}, \dd, \mathbf{d'} \ff, \mathbf{f'}} &V(T_1)_{\ee, \ff} \times \left( \prod_{i=2}^m \delta(e_i, e_i') V(T_i)_{e_i, \dd^i} \right) \\
    &\times \delta(\dd, \dd') \delta(\ff, \ff') Z(O_p)_{\dd, \ff, s}
\end{align*}
Now we will proceed to prove that this term is equal to $V(T_k) \otimes^* Z(O)$. Let $I_{T_k} \in \mathbb{S}^{e_k', \dd^k, s, e_k, \dd^k, s}$. We will calculate the value of the outside term in sections. Let $A = V(T_1) \otimes_k \left(I_{T_k}, V(T_{k+1}),\ldots ,V(T_n) \right)$. Then, 
\begin{align*}
    &A_{e_1,\ldots ,e_{k-1},\dd^k, s, \hat{e}_k, \hat{\dd}_k, \hat{s}, \dd^{k+1},\ldots , \dd^n, \ff} =\\
    &A^\pi_{e_1,\ldots ,e_{k-1},\dd^k, \dd^{k+1},\ldots , \dd^n, \ff, s, \hat{e}_k, \hat{\dd}_k, \hat{s}} = \\
    &\sum_{\substack{e_k,\ldots ,e_n \\ e'_k,\ldots ,e'_n}} V(T_1)_{\ee, \ff} \times \delta(e_k, e_k') \delta(\dd^k, \hat{\dd}^k) \delta(s, \hat{s}) \times \prod^m_{i=k+1} \delta(e_i, e_i') V(T_i)_{e_i', \dd^i} \\
    &\left[A^\pi \otimes (V(T_2),\ldots ,V(T_{k-1})) \right]_{\dd, \ff, s, \hat{e}_k, \hat{\dd}^k, \hat{s}} = \\
    &\sum_{\ee, \ee'} V(T_1)_{\ee, \ff} \times \prod_{\substack{i=2\\i \neq k}}^n V(T_i)_{e_i', \dd^i} \times \\ &\hspace{3cm}\delta(\ee, \ee') \times \delta(e_k, \hat{e}_k) \times \delta(\dd^k, \hat{\dd}^k) \times \delta(s, \hat{s}) \\
    &[Z(O)]_{\hat{e}_k, \hat{\dd}^k, \hat{s}} =
    \sum_{\substack{\ee, \ee', \dd, \dd' \\ \ff, \ff', s, s'}}V(T_1)_{\ee, \ff} \times \prod_{\substack{i=2\\i \neq k}}^n V(T_i)_{e_i', \dd^i} \times Z(O_p)_{\dd', \ff', s'} \\  &\hspace{2cm} \times \delta(\ee, \ee') \times \delta(e_k, \hat{e}_k) \times \delta(\dd^k, \hat{\dd}^k) \times \delta(s, \hat{s}) \\
    & \hspace{2cm}\times \delta(\dd, \dd') \times \delta(\ff, \ff') \times \delta(s, s') \\
    &[V(T_k) \otimes^* Z(O)]_{\hat{s}} = \\ & \hspace{1cm} \sum_{\substack{\ee, \ee', \dd, \dd' \\ \ff, \ff', s, s' \\ \hat{e}_k, \hat{\dd}^k, e''_k, \dd^{k''}}} V(T_k)_{e''_k, \dd^{k''}} \times V(T_1)_{\ee, \ff} \times \prod_{\substack{i=2\\i \neq k}}^n V(T_i)_{e_i', \dd^i} \times Z(O_p)_{\dd', \ff', s'} \\  &\hspace{2cm} \times \delta(\ee, \ee') \times \delta(e_k, \hat{e}_k) \times \delta(\dd^k, \hat{\dd}^k) \times \delta(s, \hat{s})  \\ & \hspace{2cm}\times \delta(\dd, \dd') \times \delta(\ff, \ff') \times \delta(s, s') \times \delta(e''_k, \hat{e}_k) \times \delta(\dd^{k''}, \hat{\dd}^k) \\
    & \hspace{1cm} = \sum_{\ee, \mathbf{e'}, \dd, \mathbf{d'} \ff, \mathbf{f'}} V(T_1)_{\ee, \ff} \times \prod_{i=2}^m  V(T_i)_{e_i, \dd^i}  \times Z(O_p)_{\dd, \ff, \hat{s}}\\
    &\hspace{2cm} \times \delta(\ee, \ee') \times \delta(\dd, \dd') \times \delta(\ff, \ff') 
\end{align*}
Which completes the proof. The last simplification step is
obtained by replacing $\hat{e}_k$ and $e''_k$ with $e_k$,
$\hat{\dd}^k$ and $\dd^{k''}$ with $\dd^k$ and $s$ and $s'$
with $\hat{s}$ since these need to be equal for any term to
contribute to the final sum. The commutativity of
$\mathbb{S}$ then allows $V(T_k)_{e_k, \dd^k}$ to be moved
to its place in the sequence. 

\end{proof}

\begin{manualtheorem}{6.4}\label{thm:outside}
   If $x$ is the goal item, then $Z(x) = I_s$. Else,
 \begin{align*}
     Z(x) = \bigoplus_{\substack{j,a_1,..,a_k,b \text{ s.t. } \\ \frac{a_1\ldots a_k}{b} \text{ and } x=a_j}}  (V(a_1) &\otimes_k \left[I_{a_k}, V(a_{k+1}),\ldots ,V(a_n) \right])^\pi  \\ \otimes  &\left(V(a_2),\ldots ,V(a_{k-1}) \right) \otimes^* Z(b) 
 \end{align*}
\end{manualtheorem}

\begin{proof}
by definition $Z(x) = \bigoplus_{D \in outer(x)} Z(D)$. Either $x$ is a goal item, in which case $Z(x) = Z({}) = I_S$. 

Otherwise the outer trees $outer(x)$ could be written as the union of outer trees $outer\left(k, \frac{a_1\ldots a_n}{b} \right)$ for each rule $\frac{a_1\ldots a_n}{b}$ where $a_k = x$ for some $k$. Hence:

\[Z(x) = \bigoplus_{\substack{j,a_1,..,a_k,b \text{ s.t. } \\ \frac{a_1\ldots a_k}{b} \text{ and } x=a_j}} \bigoplus_{D \in outer\left(k,\frac{a_1\ldots a_n}{b}\right)} Z(D) \]

For the inner part of this equation we have:

\begin{align*}
    &\bigoplus_{D \in outer\left(k,\frac{a_1\ldots a_n}{b}\right)} Z(D) = \\  &\hspace{1cm}\bigoplus_{D_b \in outer(b)} \bigoplus_{\substack{D_{a_1} \in inner(a_1),\ldots , \\
                         D_{a_{k-1}}\in inner(a_{k-1})}} \bigoplus_{\substack{D_{a_{k+1}} \in inner(a_{k+1}),\ldots , \\
                         D_{a_n} \in inner(a_n)}} \\
    & \hspace{2cm} \left( V(D_{a_1}) \otimes_k \left[ I_{D_{a_k} \times d_S}, V(D_{a_{k+1}}),\ldots ,V(D_{a_n}) \right] \right)^\pi \\
    & \hspace{2cm} \otimes \left( V(D_{a_2}),\ldots ,V(D_{a_{k-1}}) \right) \otimes^* Z(D_b)     
\end{align*}
Since $\oplus$ distributes over $\otimes$, this can rewritten as
\begin{align*}
    &\bigoplus_{D \in outer\left(k,\frac{a_1 \ldots a_n}{b}\right)} Z(D) = \\  
    &\left( 
        \bigoplus_{
            \substack{
                D_{a_1} \in \\ 
                inner(a_1)
            }
        } 
        V(D_{a_1}) \otimes_k 
        \left[ 
            I_{D_{a_k}}, 
            \bigoplus_{
                \substack{
                    D_{a_{k+1}} \in \\ 
                    inner(a_{k+1})
                }
            } 
            V\left(D_{a_{k+1}}\right), \ldots,  
            \bigoplus_{
                \substack{D_{a_n} \in \\ 
                inner(a_n)
                }
            } 
            V\left(D_{a_n}\right) 
        \right] 
    \right)^\pi \\
    &{\hspace{1cm} \otimes \left( \bigoplus_{D_{a_2} \in inner(a_2)} V(D_{a_2}),\ldots, \bigoplus_{D_{a_{k-1}} \in inner(a_{k-1})} V(D_{a_{k-1}}) \right)} \\  &\hspace{1cm}\otimes^*  \bigoplus_{D_b \in outer(b)} Z(D_b)
\end{align*}

And since $V(a_i)$ and $Z(D_b)$ are defined as the summation of their inner and outer trees respectively 

\begin{align*}
    &\bigoplus_{D \in outer\left(k,\frac{a_1\ldots a_n}{b}\right)} Z(D) = \\  &\hspace{1cm} \left(V(a_1) \otimes_k \left[I_{a_k}, V(a_{k+1}),\ldots ,V(a_n) \right]\right)^\pi \otimes  \left(V(a_2),\ldots ,V(a_{k-1}) \right) \otimes^* Z(b) 
\end{align*}

Replacing the inner part of the previous equation with this term gives us the desired equality, completing the proof.
\end{proof}

\section*{Appendix B - Inside and Outside Calculations for Looping Buckets}

In computing the inside and outside values with an item-based description, we assume a pre-computed ordering over items in the form of \textit{buckets}. For items $x$ and $y$, we write $bucket(x) \leq bucket(y)$ if the value of $y$ depends on the value of $x$. So far we have assumed that items could be simply sorted so that no item directly or indirectly depends on itself, and given the inside and outside formulas accordingly. In this section we give the equivalent formulas for items in \textit{looping buckets}. Items in a looping bucket depend on each other and computing their values might require an infinite sum. Our presentation and proofs both follow that of \citet{Goodman1998ParsingInside-Out}.

For an item $x$ in a looping bucket $B$, let the \textit{generation} of a derivation tree $x$ to be the maximum number of items in $B$ that could appear in a single path from the root to a leaf. This intuitively provides an ordering for processing a potentially infinite number of trees by starting from generation 0 and incrementally adding larger and larger trees. We will denote the set of inner trees of $x$ with generation at most $g$ with $inner_{\leq}(x,B)$ Adding up the values of all inner trees of $x$ that have generation at most $g$ then gives us an approximation for the true inner value of $x$, and the approximation gets better as $g$ gets larger. Formally, we define a \textit{g generation value} for an item $x$ in bucket $B$ as:
\[V_{\leq g}(x,B) = \bigoplus_{D \in inner_{\leq g}(x,B)} V(D) \]
For $\omega$-continuous semirings, the infinite sum is equal to the supremum of the partial sums (\citealt{Kuich1997SemiringsAutomata}, 613), hence (\citealt{Goodman1999SemiringParsing}, 589):
\[V(x) = \bigoplus_{D \in inner(x)} V(D) = \sup_g V_{\leq g}(x,B)\]

Fortunately, tensors of semirings of set dimensions are $\omega$-continuous as long as the underlying semiring is $\omega$-continuous. We give the necessary definitions to establish this property:

\begin{definition}(\citealt{Kuich1997SemiringsAutomata}, 611)
A semiring is \textbf{naturally ordered} if there is a partial ordering $\sqsubseteq$ such that $x \sqsubseteq y$ iff there is a $z$ s.t. $x \oplus z = y$.
\end{definition}

\begin{definition} (\citealt{Kuich1997SemiringsAutomata}, 612)
A naturally ordered complete semiring is $\omega$-continuous if for any sequence $x_1,x_2,\ldots$ and for any constant $y$, if for all $n$, $\bigoplus_{0 \leq i \leq n} x_i \sqsubseteq y$ then $\bigoplus_i x_i \sqsubseteq y$
\end{definition}

Notice that for the set of tensors in $\S^\mathbf{d}$ where $\mathbf{d}$ is an arbitrary list of positive integers, if the underlying semiring has a natural ordering then this could be extended straightforwardly to $\S^\mathbf{d}$ by the following rule: $\mathbf{X} \sqsubseteq \mathbf{Y}$ iff $\mathbf{X}_\mathbf{i} \sqsubseteq \mathbf{Y}_\mathbf{i}$ for all indices $\mathbf{i}$. It is straightforward to check that if the underlying semiring is $\omega$-continuous, then $\S^\mathbf{d}$ is $\omega$-continuous as well.

\citet{Goodman1999SemiringParsing} gives a formula for $V_{\leq g} (x, B)$ in order to compute or approximate the supremum. Below we give the analogous formula for partial semirings:

\begin{manualtheorem}{B.1}
For items $x$ in a looping bucket $B$ and the generation $g \geq 1$ 

\[V_{\leq g}(x,B) =  \bigoplus_{\substack{[a_1,\ldots ,a_k] \\ \text{s.t.}\frac{a_1,..,a_k}{x}}} K_g(a_1, B) \otimes \left[  K_g(a_2, B) , \ldots ,K_g(a_k, B) \right] \] 
Where
\[ K_g(a, B) =  
\begin{cases}
    V(a) &\text{ if } a \notin B \\
    V_{\leq g-1}(a, B) &\text{ if } a \in B
\end{cases}
\]
\end{manualtheorem}

\begin{proof}
\begin{align*}
    V_{\leq g}(x, B) &= \bigoplus_{D \in inner_{\leq g}(x, B)} V(D) \\
    &= \bigoplus_{\substack{[a_1,\ldots ,a_k] \\ \text{s.t.}\frac{a_1,..,a_k}{x}}} 
    \bigoplus_{\substack{D_{a_1} \in inner_{\leq g-1}(a_1, B),\ldots, \\
                         D_{a_k}\in inner_{\leq g-1}(a_k, B)}} 
    V \left( \langle x: D_{a_1}, \ldots D_{a_k} \rangle \right) \\
    &= \bigoplus_{\substack{[a_1,\ldots,a_k] \\ \text{s.t.}\frac{a_1,..,a_k}{x}}} 
    \bigoplus_{\substack{D_{a_1} \in inner_{\leq g-1}(a_1, B),\ldots, \\
                         D_{a_k}\in inner_{\leq g-1}(a_k, B)}}
    V(D_{a_1}) \otimes [V(D_{a_2}), \ldots, V(D_{a_k})] \\
    &= \bigoplus_{\substack{[a_1,\ldots,a_k] \\ \text{s.t.}\frac{a_1,..,a_k}{x}}} \bigoplus_{D_{a_1} \in inner_{\leq g-1}(a_1, B)} V(D_{a_1}) \\ &\hspace{2cm}\otimes \left[ \bigoplus_{D_{a_2} \in inner_{\leq g-1}(a_2, B)} V(D_{a_2}), \ldots, \bigoplus_{D_{a_k} \in inner_{\leq g-1}(a_k, B)} V(D_{a_k}) \right] \\
    &= \bigoplus_{\substack{[a_1,\ldots,a_k] \\ \text{s.t.}\frac{a_1,..,a_k}{x}}} V_{\leq g-1}(a_1, B) \otimes \left[V_{\leq g-1}(a_2, B), \ldots, V_{\leq g-1}(a_k, B) \right]
\end{align*}

Note that if $a_i$ is not in the bucket $B$ then $V_{\leq g-1}(a_i, B) = V(a_i)$, hence $V_{\leq g-1}(a_i, B)$ can be replaced with $K_g(a_i, B)$, completing the proof. 
\end{proof}

We will follow a similar strategy for computing the outside values of items that belong to a looping bucket. The only difference is the slight difference in the definition of the generation of of the tree. If $D \in outer(x)$ where $x$ belongs to a looping bucket $B$, then the generation of $D$ is maximum number of items that could appear in a single path from the root to $x$, where $x$ is included in the count. Let 

\[Z_{\leq g}(x, B) = \bigoplus_{D \in outer_{\leq g}(x, B)} Z(D)\]

\begin{manualtheorem}{B.2}
For items $x$ in a looping bucket $B$ and the generation $g \geq 1$

\begin{align*}
     Z_{\leq g}(x, B) =  \bigoplus_{\substack{j,a_1,..,a_k,b \text{ s.t. } \\ \frac{a_1\ldots a_k}{b} \text{ and } x=a_j}} \hspace{-4mm}  (V(a_1) &\otimes_k \left[I_{a_k}, V(a_{k+1}),\ldots,V(a_n) \right])^\pi \\
     \otimes  &\left[(V(a_1),\ldots,V(a_{k-1}) \right] \otimes^* H_g(b, B) 
 \end{align*} 
 Where $\pi$ is defined as in Theorem \ref{thm:outside} and
 \[ H_g(b, B) =  
\begin{cases}
    Z(b) &\text{ if } b \notin B \\
    Z_{\leq g-1}(b, B) &\text{ if } b \in B
\end{cases}
 \]
\end{manualtheorem}

\begin{proof}
\begin{align*}
    Z_{\leq g}(x, B) &= \bigoplus_{D \in outer_{\leq g}(x, B)} Z(D) \\
    &=
    \bigoplus_{
        \substack{
            j,a_1,..,a_k,b \text{ s.t. } \\ 
            \frac{a_1\ldots a_k}{b} \text{ and } x=a_j
        }
    } 
    \bigoplus_{ 
        D \in outer_{\leq g-1} 
        \left(
            k, \frac{a_1\ldots a_n}{b}
        \right)
    } Z(D) \\
    &= 
    \bigoplus_{
        \substack{j,a_1,..,a_k,b \text{ s.t. } \\ 
            \frac{a_1\ldots a_k}{b} \text{ and } x=a_j
        }
    } 
    \bigoplus_{ D_b \in outer_{\leq g-1}(b) } 
    \bigoplus_{
        \substack{
            D_{a_1} \in inner(a_1),\ldots , \\ 
            D_{a_{k-1}} \in inner(a_{k-1})
        }
    } 
    \bigoplus_{
        \substack{
            D_{a_{k+1}} \in inner(a_{k+1}),\ldots , \\  
            D_{a_n} \in inner(a_n)
        }
    } \\
    & \hspace{2cm} \left( V(D_{a_1}) \otimes_k \left[ I_{D_{a_k} \times d_S}, V(D_{a_{k+1}}),\ldots ,V(D_{a_n}) \right] \right)^\pi \\
    & \hspace{2cm} \otimes \left( V(D_{a_2}),\ldots ,V(D_{a_{k-1}}) \right) \otimes^* Z_{\leq g}(D_b, B) \\
    &= \hspace{-3mm}
    \bigoplus_{
        \substack{
            j,a_1,..,a_k,b \text{ s.t. } \\ 
            \frac{a_1\ldots a_k}{b} \text{ and } x=a_j
        }
    }  
    \left( 
        \bigoplus_{
            \substack{
                D_{a_1} \in \\ 
                inner(a_1)
            }
        } 
        V(D_{a_1}) \otimes_k 
        \left[ 
            I_{D_{a_k}}, \hspace{-3mm}
            \bigoplus_{
                \substack{
                    D_{a_{k+1}} \in  \\ 
                    inner(a_{k+1})
                }
            } \hspace{-3mm} V(D_{a_{k+1}}),\ldots, \hspace{-3mm} 
            \bigoplus_{
                \substack{
                    D_{a_n} \in \\ 
                    inner(a_n)
                }
            } \hspace{-3mm} V(D_{a_n}) 
        \right] 
    \right)^\pi \\
    &\hspace{1cm} \otimes \left( \bigoplus_{D_{a_2} \in inner(a_2)} V(D_{a_2}),\ldots , \bigoplus_{D_{a_{k-1}} \in inner(a_{k-1})} V(D_{a_{k-1}}) \right) \\  &\hspace{1cm}\otimes^*  \bigoplus_{D_b \in outer_{\leq g-1}(b)} Z_{\leq g-1}(D_b, B)  \\
    &=\bigoplus_{\substack{j,a_1,..,a_k,b \text{ s.t. } \\ \frac{a_1\ldots a_k}{b} \text{ and } x=a_j}} (V(a_1) \otimes_k \left[I_{a_k}, V(a_{k+1}),\ldots ,V(a_n) \right])^\pi  \\ &\hspace{5cm} \otimes  \left[(V(a_2),\ldots ,V(a_{k-1}) \right] \otimes^* Z_{\leq g-1}(b, B)
\end{align*}

Like the inner case, note that for an item $b$ not in the looping bucket $b$, $Z_{\leq g-1}(b,B) = Z(b)$, hence we can replace $Z_{\leq g-1}(b,B)$ with $H_g(b,B)$, completing the proof.
\end{proof}

\end{document}